\documentclass[a4paper,conference]{IEEEtran}

\ifCLASSINFOpdf

\else

\fi

\usepackage{url}
\usepackage{bm}
\usepackage{mathrsfs}

\usepackage{tablefootnote}
\usepackage{amsmath}
\usepackage{amsfonts}
\usepackage{graphicx}
\usepackage{subfigure}
\usepackage{flushend}
\usepackage{epstopdf}
\usepackage{threeparttable}
\usepackage{epsfig}
\usepackage{url}
\usepackage{balance}
\usepackage{array}
\usepackage{booktabs}
\usepackage{flushend}
\usepackage{float}
\usepackage{cite}
\usepackage[misc]{ifsym}
\usepackage{algorithm}
\usepackage{algorithmic}
\usepackage{hyperref}
\hypersetup{
    colorlinks=true,
    linkcolor=blue,
    filecolor=magenta,
    urlcolor=cyan,
}

\def\ie{\emph{i.e.}}
\def\eg{\emph{e.g.}}

\begin{document}

\def\x{{\mathbf x}}
\def\L{{\cal L}}

\title{Self-Paced Deep Regression Forests\\
for Facial Age Estimation
}

\author{\IEEEauthorblockN{Shijie Ai\IEEEauthorrefmark{2},
\Letter~{Lili Pan}\footnote{Corresponding author.}\IEEEauthorrefmark{2} and
Yazhou Ren\IEEEauthorrefmark{4}}
\IEEEauthorblockA{\IEEEauthorrefmark{2}School of Information and Communication Engineering}
\IEEEauthorblockA{\IEEEauthorrefmark{4}School of Computer Science and Engineering\\
University of Electronic Science and Technology of China, Chengdu 611731, China\\
Email: shijieai@std.uestc.edu.cn, \{lilipan, yazhou.ren\}@uestc.edu.cn }
}

\maketitle

\begin{abstract}
Facial age estimation is an important and challenging problem in computer vision.
Existing approaches usually employ deep neural networks (DNNs) to fit the mapping from facial features to age, even though there exist some noisy and confusing samples.
We argue that it is more desirable to distinguish noisy and confusing facial images from regular ones, and alleviate the interference arising from them.
To this end, we propose self-paced deep regression forests (SP-DRFs) -- a gradual learning DNNs framework for age estimation.
As the model is learned gradually, from simplicity to complexity, it tends to emphasize more on reliable samples and avoid bad local minima.
Moreover, the proposed capped-likelihood function helps to exclude noisy samples in training, rendering our SP-DRFs more robust.
We demonstrate the efficacy of SP-DRFs on Morph II and FG-NET datasets, where our model shows significantly improved performance when compared with non-gradual learning age estimation models.
\end{abstract}


%
\IEEEpeerreviewmaketitle

\section{Introduction}
\label{sec:intro}
Facial age estimation~\cite{gunay2008automatic,guo_human_2009,Yu2010Multi,huang_soft-margin_2017,niu_ordinal_2016,chen_using_2017,gao_age_2018,shen_deep_2018,li2019bridgenet} involves the learning of the mapping from facial features to age, such that given a new face image, its corresponding age can be predicted.
Due to its wide potential applications, ranging from human computer interaction (HCI) to age based advertising, numerous research studies have been devoted to this research field, which can be mainly divided into shallow model based or DNNs based.

Shallow model based approaches, for example~\cite{gunay2008automatic,guo_human_2009,Yu2010Multi,huang_soft-margin_2017}, model the nonlinear mapping from facial features to age using traditional classifier or regressor.
Amongst them, classification based approaches categorize real age into independent groups and learn a classifier, such as $k$-nearest neighbors (KNN)~\cite{gunay2008automatic} or support vector machine (SVM)~\cite{guo_human_2009}, to classify the age.
However, this kind of approach can only be used to estimate age categories but not continuous age.
Besides, it assumes the age groups are independent and have no inherent
relationship to each other, and hence may omit the inter-relationship (ordinal information) among age groups.
Regression based approaches, such as Gaussian process regression~\cite{Yu2010Multi}, support vector regression (SVR)~\cite{guo_human_2009} and soft-margin mixture of regressions~\cite{huang_soft-margin_2017}, usually learn a nonlinear mapping function between facial features and age, but are prone to learn a biased mapping due to the unbalanced data and limited model capacity.

DNNs based approaches, for example~\cite{niu_ordinal_2016,chen_using_2017,gao_age_2018,shen_deep_2018, li2019bridgenet}, employ DNNs to model the age mapping more precisely.
Ordinal-based approaches~\cite{niu_ordinal_2016,chen_using_2017} resort to a set of sequential binary queries -- each query refers to a comparison with a predefined age, to exploit the inter-relationship (ordinal information) among age labels. 
Furthermore, improved deep label distribution learning (DLDL)~\cite{gao2017deep} explores the underlying age distribution patterns to effectively accommodates age ambiguity.
Besides, deep regression forests (DRFs)~\cite{shen_deep_2018} connect random forests to deep neural networks and achieve promising results.
BridgeNet~\cite{li2019bridgenet} uses local regressors to partition the data space and gating networks to provide continuity-aware weights.
The final age estimation result is the mixture of the weighted regression results.
Overall, these DNNs based approaches have enhanced age estimation performance largely; however, they plausibly ignore one problem: the interference arising from confusing and noisy samples -- facial images with PIE (\ie~pose, illumination and expression) variation, occlusion, misalignment and so forth.
For this reason, existing DNNs based methods lack robustness.

\begin{figure*}
\centerline{\includegraphics[width=\textwidth]{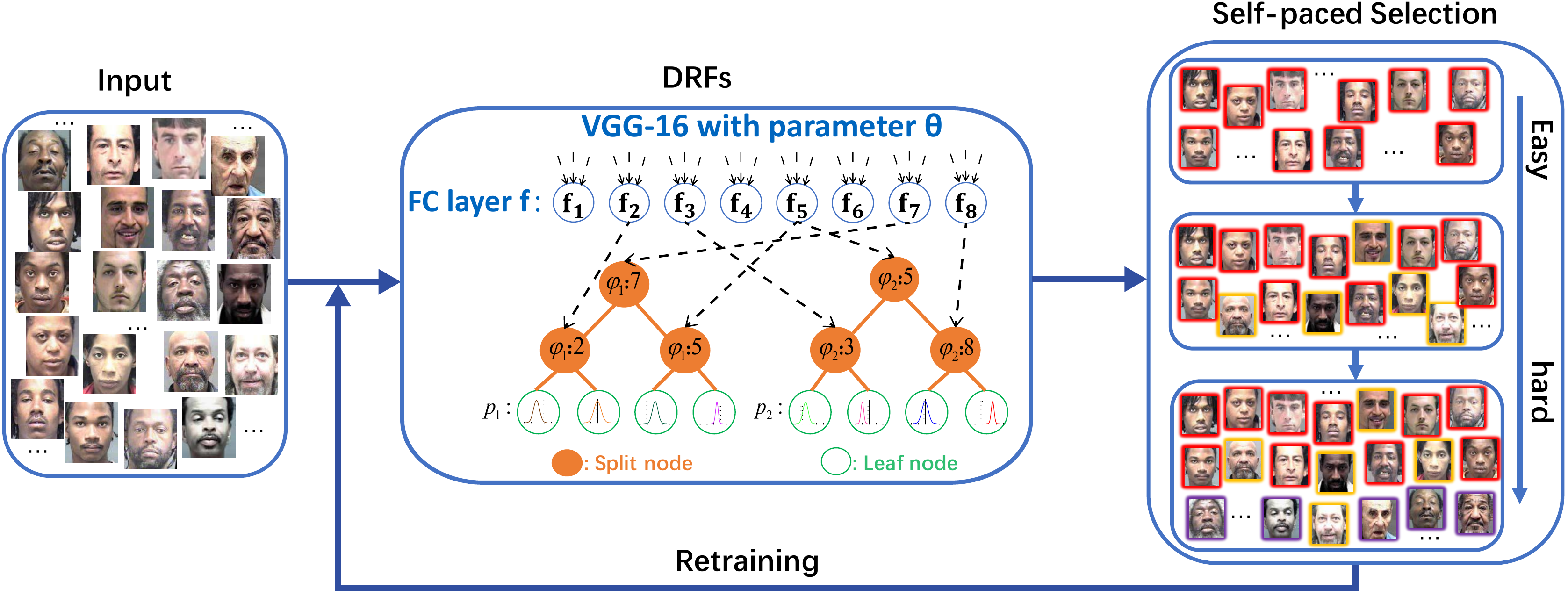}}
\caption{Illustration of our proposed self-paced deep regression forests (SP-DRFs). All input images are ranked according to their reliability, which is estimated according to the likelihood of the DRFs parameters given each $\left\{\mathbf{x}_i, t_i\right\}$, \ie~$p_{\mathcal{F}}\left(t_i|\mathbf{x}_i,\bm{\Theta},\bm{\Pi} \right)$, where the likelihood is obtained based on the estimated $\left\{\bm{\Theta},\bm{\Pi}\right\}$ in the previous pace. Under each ranking, we progressive select more hard facial images to retain our model.
In other words, different subsets of the facial images are selected, from easy to hard gradually, for retraining. Such a process is just like the learning process of our human beings.}
\label{Fig1}
\end{figure*}

Thus, one crucial, yet elusive problem in DNNs based age estimation methods is: \emph{how to alleviate the influence of confusing and noisy facial images during training?}

This problem has been studied in recent works~\cite{zeng2019soft,guodong_guo_image-based_2008} as how to learn robust and discriminative facial features through a rather deep neural network with a regularized loss function.
However, they do not truly distinguish the confusing and noisy samples from regular ones, and yet not exclude the interference caused by the unreliable samples, resulting in unsatisfactory robustness.
Our work, from another perspective, mimic the cognitive proposes of our brain -- from simplicity to complexity and propose self-paced deep regression forests (SP-DRFs) -- a gradual learning algorithm for DNNs based age estimation.
Owing to the gradual learning strategy and the proposed capped-likelihood mechanism, our SP-DRFs emphasize more on ``good'' samples and have the capability to exclude noisy samples, resulting in improved robustness, where bad local minima are avoided as well.

We demonstrate the efficacy of SP-DRFs on the Morph II and FG-NET datasets, where our method is shown to achieve state-of-the-art performance.
Especially, our method attains a value of 1.98 for mean absolute error on the Morph \uppercase\expandafter{\romannumeral2} dataset, which obviously outperforms one of the previous state-of-the-art method -- DRFs, and obtain an improvement of 8.7\%.
This demonstrates the validity of our viewpoint: learning an age predictive model from simplicity to complexity -- like our human beings, can improve its robustness and avoid bad local minima to some extent.

\section{Deep Regression Forests and Notations}

In this section, we review the basic concepts of deep regression forests (DRFs)~\cite{shen_deep_2018}.

\noindent \textbf{Deep regression tree.} DRFs generally consist of a certain number of deep regression trees.
Each deep regression tree, given a training set of input-output pairs $\left\{\mathbf{x}_i, t_i\right\}_{n=1}^N$, learns DNNs coupled with a regression tree $\mathcal{T}$ that map input images to target output.
Typically, a regression tree $\mathcal{T}$ contains two kinds of nodes: split (or decision) nodes and leaf (or prediction) nodes~\cite{shen_deep_2018} (see Fig.~\ref{Fig1}).
More specifically, the split node determine a sample goes to the left or right sub-tree; each leaf node determine the predicted age distribution of each sample.


\noindent \textbf{Split node.}
Each split node corresponds to a split function, $s_{n}(\mathbf{x} ; \bm{\Theta}) : \mathbf{x} \rightarrow[0,1]$, where $\bm{\Theta}$ is the parameters of DNNs.
Conventionally, the split function is formulated as $s_{n}(\mathbf{x} ; \bm{\Theta})=\sigma\left(\mathbf{f}_{\varphi(n)}(\mathbf{x}; \bm{\Theta})\right)$, where $\sigma(\cdot)$ is the sigmoid function, $\mathbf{f}(\mathbf{x}; \bm{\Theta})$ is the learned deep features through DNNs given an input $\mathbf{x}$, and $\varphi(\cdot)$ is an index function to specify the split node $n$ that the $\varphi(n)$-th element of $\mathbf{f}(\mathbf{x}; \bm{\Theta})$ is associated with.
Fig.~\ref{Fig1} plots an example DRFs model with two trees, where $\varphi_1$ and $\varphi_2$ are the index functions.
Under the split rules in DRFs, the probability of the sample $\mathbf{x}_i$ falling into the leaf node $\ell$ is given by:

\begin{equation}
\label{Eq.1}
\omega_\ell( \mathbf{x} | \bm{\Theta)}=\prod_{n \in \mathcal{N}} s_{n}(\mathbf{x}; \Theta)^{[\ell \in \mathcal{L}_{n_l}]}\left(1-s_{n}(\mathbf{x}; \Theta)\right)^{\left[\ell \in \mathcal{L}_{n_r}\right]},
\end{equation}
where $l \in \mathcal{L}_{n_l}$ denotes the the sets of leaf nodes owned by the sub-tree $\mathcal{T}_{n_l}$, and $l \in \mathcal{L}_{n_r}$ denotes the leaf nodes owned by the sub-tree $\mathcal{T}_{n_r}$.
$\mathcal{T}_{n_l}$ and $\mathcal{T}_{n_r}$ are rooted at the left and right children ${n}_{l}$ and ${n}_{r}$ of node $n$, respectively.
$[l \in \mathcal{L}_{n_l}]$ is an indicator function, having the value 1 for all nodes $n$ those are satisfy $l \in \mathcal{L}_{n_l}$ and the value 0 for all nodes $n$ those are not satisfy $[l \in \mathcal{L}_{n_l}]$.
$[l \in \mathcal{L}_{n_r}]$ is defined similarly.

\noindent \textbf{Leaf node.} For tree $\mathcal{T}$, given $\mathbf{x}$,  each leaf node $\ell \in \mathcal{L}$ defines a distribution of predicted age $t$, denoted by $p_{\ell}(t)$.
To be specific, $p_{\ell}(t)$ is assumed to be a Gaussian distribution with mean $\mu_l$, and variance $\sigma^2_l$.
Thus, considering all leaf nodes, the final distribution of $t$ conditioned on $\mathbf{x}$ is averaged by the probability of reaching each leaf:
\begin{equation}
\label{Eq.2}
p_{\mathcal{T}}(t | \mathbf{x} ; \bm{\Theta}, \bm{\pi})=\sum_{\ell \in \mathcal{L}} \omega_\ell( \mathbf{x} | \bm{\Theta)} p_{\ell}(t),
\end{equation}
where $\bm{\Theta}$ and $\bm{\pi}$ represent the parameters of DNNs and the distribution parameters $\left\{\mu_l, \sigma^2_l, l \in \mathcal{L}\right\}$, respectively.
It can be viewed as a mixture distribution, where $\omega_\ell( \mathbf{x} | \bm{\Theta)}$ denotes mixing coefficients and $ p_{\ell}(t)$ denotes sub-modal distributions.
One note that the distribution parameters vary along with tree $\mathcal{T}_k$, and thus we rewrite them in terms of $\bm{\pi}_k$.

\noindent \textbf{Forests of regression trees.}
As previously mentioned, a forest comprises a set of deep regression trees, \ie, $\mathcal{F}=\left\{\mathcal{T}_1,...,\mathcal{T}_k\right\}$.
Hence, the final predictive age distribution $p_{\mathcal{F}}\left(t|\mathbf{x},\bm{\Theta},\bm{\Pi} \right)$ should be formulated as the average distribution of the predicted age calculated from all trees:
\begin{equation}
\label{Eq.3}
p_{\mathcal{F}}\left(t|\mathbf{x},\bm{\Theta},\bm{\Pi} \right)
=
\frac{1}{K}\sum_{k=1}^K p_{\mathcal{T}_k}\left(t|\mathbf{x}; \bm{\Theta}, \bm{\pi}_k\right),
\end{equation}
where $K$ is the number of trees and $\bm{\Pi}=\left\{\bm{\pi}_1,...,\bm{\pi}_K\right\}$.
K trees setting alleviates the randomness of the predicted result calculated from only one tree.
Moreover, K trees share the same DNNs, which will not dramatically increase the number of parameters in the DRFs.

\section{Self-paced learning protocol}
\label{Sec.3}

\subsection{Self-Paced Learning}
Self-paced learning is a gradual learning paradigm, which imitates human cognitive process.
It is built on the intuition that, rather than considering all training samples simultaneously, the algorithm should be presented with the training data from easy to difficult, which facilitates learning~\cite{Kumar2010Self, meng_theoretical_2017}.
Typically, a self-paced regularization term is incorporated into the objective, rendering faster convergence rate, and better optimized solution -- more robust to confusing and noisy data~\cite{Kumar2010Self, meng_theoretical_2017}.

\subsection{Proposed Formulation}
To promote robustness of age estimation, unlike existing DNNs based approaches attempting to learn discriminative features~\cite{niu_ordinal_2016, chen_using_2017, gao_age_2018, shen_deep_2018, li2019bridgenet}, we directly distinguish noisy and confusing facial images from regular ones, and alleviate the interference arising from them.
To this end, we propose self-paced deep regression forests (SP-DRFs), which provides a way to gradually learn DRFs, from simplicity to complexity, until the learned model is ``mature".
For this purpose, we shall define a latent variable $v_i$ that indicates whether the $i^{th}$ sample is selected ($v_i=1$) or not ($v_i=0$) depending on how reliable it is for training.
In age estimation scenario, easy or reliable samples mean regular face images, probably without noise, occlusion, PIE variations, misalignment and so on.
Thus, our target is to jointly maximizing the log likelihood function with respect to DRFs' parameters $\bm{\Pi}$ and $\bm{\Theta}$, and learn the latent selecting variables $\mathbf{v}=\left[v_1,...,v_N\right]^T$:
\begin{equation}
\label{Eq.4}
 \max_{\bm{\Theta},\bm{\Pi}, \mathbf{v}} \sum_{i=1}^{N} v_{i} \log p_{\mathcal{F}}\left(t_i|\mathbf{x}_i,\bm{\Theta},\bm{\Pi} \right)   + \lambda\sum_{i=1}^N v_i,
\end{equation}

\noindent where $\lambda>0$, and whether $\log p_{\mathcal{F}}\left(t_i|\mathbf{x}_i,\bm{\Theta},\bm{\Pi} \right) + \lambda > 0$ determines the $i^{th}$ sample is selected or not.
If $\lambda$ is small, maximizing Eq.~(\ref{Eq.4}) would only involve easy samples, for which the likelihoods of having correctly predicted age are high; if $\lambda$ is large,  maximizing Eq.~(\ref{Eq.4}) would involve more unreliable samples, whose likelihoods are low.

Iteratively increasing $\lambda$, samples are dynamically involved in the training of DRFs, starting with easy samples and ending up with all samples.
Note every time we retrain DRFs, that is, maximizing Eq.~(\ref{Eq.4}), our model is initialized to the results of last iteration.
As such, our model is initialized progressively more reasonably in each retraining iteration -- adaptively calibrated by selected reliable samples.
This also means we place more emphasis on reliable facial images rather than confusing and noisy ones.
Following this line of reasoning, SP-DRFs are prone to have significantly more robust solution and can avoid bad local minima to some extent.
Despite recent works~\cite{Li2016Self, ijcai2017, jiang2014self, ma2017self, lin2017active, ren2018self, ren2019self} have investigated the benefits of self-paced DNNs, but our work is the first to demonstrate its efficacy on facial age estimation application.

In addition, one note the above SP-DRFs model is not capable of excluding noisy samples, only places more emphasis on reliable facial images.
To fix this drawback, we propose to construct capped-likelihood function in our SP-DRFs to solve for model parameters, which is inspired by recent self-paced learning work~\cite{ren2018self}.
Specifically, the capped function render the sample likelihood with especially small value, to zero:
\begin{equation}
\label{eq:cap}
\text{cap}\left(p, \epsilon\right) = \frac{\max\left(p-\epsilon, 0\right)}{p-\epsilon}p.
\end{equation}
Here, $p$ denotes the likelihood, and $\epsilon$ denotes the threshold ($\epsilon>0$).
Given the sample likelihood $p_{\mathcal{F}}\left(t_i|\mathbf{x}_i,\bm{\Theta},\bm{\Pi} \right)$, the capped likelihood is denoted as
$ p^c_{\mathcal{F}}\left(t_i|\mathbf{x}_i,\bm{\Theta},\bm{\Pi} \right)$.
As the log likelihoods of the noisy samples are prone to have especially small values, such a capping operation sets them to be $-\infty$.
Thus, incorporating capped-likelihoods into the objective function potentially avoid the interference of noisy samples by virtue of learned selecting variables:
\begin{equation}
\label{Eq.5}
 \max_{\bm{\Theta},\bm{\Pi}, \mathbf{v}} \sum_{i=1}^{N} v_{i} \log p^c_{\mathcal{F}}\left(t_i|\mathbf{x}_i,\bm{\Theta},\bm{\Pi} \right)   + \lambda\sum_{i=1}^N v_i.
\end{equation}
Proposing to construct the capped-likelihood function in our SP-DRFs is one of the main contributions of this paper.

\subsection{Learning by Alternative Search}
\label{Learning}

\begin{figure*}[t]
\centerline{\includegraphics[width=\textwidth]{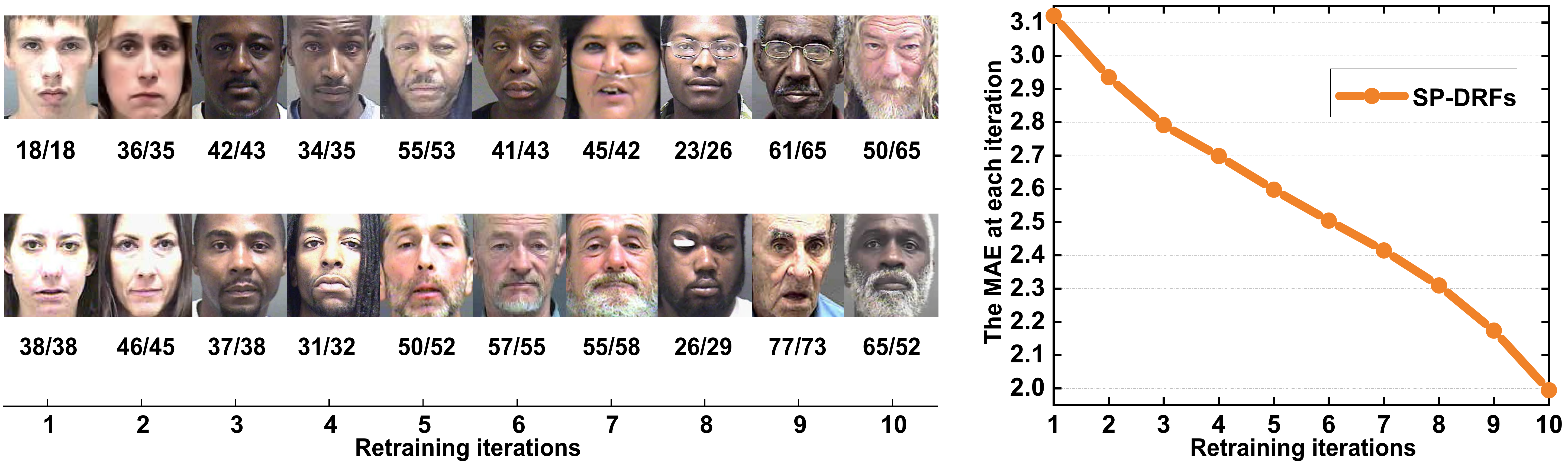}}
\caption{The gradual learning process. \textbf{Left}: The typical worst cases at each pace become more confusing and noisy along with iteratively increase $\lambda$. The two numbers below each image are the real age (left) and predicted age (right). \textbf{Right}: The MAE of SP-DRFs at each pace descends step by step. This partially demonstrates the validity of the gradual learning strategy in our DRFs.}
\label{Fig2}
\end{figure*}

The learning involves a two step alternative search strategy (ASS): (\romannumeral1) update $\mathbf{v}$ for sample selection, and (\romannumeral2) update $\bm{\Theta}$ and $\bm{\Pi}$ with current sample selection.

\renewcommand{\algorithmicrequire}{ \textbf{Input:}}
\renewcommand{\algorithmicensure}{ \textbf{Output:}}
\begin{algorithm}[b]
\caption{The training process of SP-DRFs.}
\label{alg:The}
\begin{algorithmic}[1]
\REQUIRE
$\mathcal{D}=\left\{\mathbf{x}_i, t_i\right\}_{i=1}^N$, $\lambda^0$, and $\epsilon$.
\ENSURE
Model parameters $\bm{\Pi}$ and $\bm{\Theta}$.
\STATE Initialize $\bm{\Pi}^{0}$, $\bm{\Theta}^{0}$ through training DRFs on all samples.
\STATE \textbf{while} not converged \textbf{do}
\STATE \quad \quad \textbf{while} not converged \textbf{do}
\STATE \quad \quad \quad \quad Randomly select a mini-batch from $\mathcal{D}$.
\STATE \quad \quad \quad \quad Update $\mathbf{v}$ by Eq.~(\ref{eq:Learning1}) or Eq.~(\ref{eq:Learning2}).
\STATE \quad \quad \quad \quad Update $\bm{\Theta}$ and $\bm{\Pi}$ following~\cite{shen_deep_2018}.
\STATE \quad \quad \textbf{end while}
\STATE \quad \quad Increase $\lambda$.
\STATE \textbf{end while}
\end{algorithmic}
\label{alg1}
\end{algorithm}

\noindent\textbf{Learning $\mathbf{v}$.} As previously mentioned, $v_i$ is a binary variable that indicates whether the $i^{th}$ sample is selected or not during training.
One note alternatively optimizing Eq.~(\ref{Eq.5}) is intractable due to the $\max\left(\cdot\right)$ function.
Here, we propose to learn $\mathbf{v}$ according to the current capped-likelihood values, which can render the subsequent optimization with respect to $\bm{\Pi}$ and $\bm{\Theta}$ tractable.
Specifically, with fixed $\bm{\Pi}$ and $\bm{\Theta}$, the optimal value of $v_i$ is calculated as follows.

\noindent \textbf{Case 1:}  if $p_{\mathcal{F}}\left(t_i|\mathbf{x}_i,\bm{\Theta},\bm{\Pi} \right)> \epsilon$, we obtain:
\begin{equation}
\label{eq:Learning1}
v_{i}=\left\{\begin{array}{ll}{1,} & {\log p_{\mathcal{F}}\left(t_i|\mathbf{x}_i,\bm{\Theta},\bm{\Pi} \right) + \lambda >0} \\ {0,} & {\text {otherwise }}\end{array}\right.,
\end{equation}
which is as the same as in the original ASS algorithm~\cite{Kumar2010Self}.
Here, the face images, for which the log likelihoods of having correctly predicted age satisfying $\log p_{\mathcal{F}}\left(t_i|\mathbf{x}_i,\bm{\Theta},\bm{\Pi} \right) + \lambda >0$, are selected.
Meanwhile, the capped likelihood $p^c_{\mathcal{F}}\left(t_i|\mathbf{x}_i,\bm{\Theta},\bm{\Pi} \right) = p_{\mathcal{F}}\left(t_i|\mathbf{x}_i,\bm{\Theta},\bm{\Pi} \right)$.

\noindent \textbf{Case 2:} if $p_{\mathcal{F}}\left(t_i|\mathbf{x}_i,\bm{\Theta},\bm{\Pi} \right) \leq \epsilon $, we have the capped likelihood $p^c_{\mathcal{F}}\left(t_i|\mathbf{x}_i,\bm{\Theta},\bm{\Pi} \right) = 0$,~\ie~$\log p^c_{\mathcal{F}}\left(t_i|\mathbf{x}_i,\bm{\Theta},\bm{\Pi} \right) = -\infty$. Since $\lambda$ is a positive factor, maximizing Eq.~(\ref{Eq.5}) must yields:
\begin{equation}
\label{eq:Learning2}
v_i = 0,
\end{equation}
meaning the noisy samples that have especially low likelihood values are not selected.

The parameter $\lambda$ could be initialized to obtain 10\% samples to train the model, then progressively increase to involve 10\% more data in each iteration.
The training stops when all the samples are involved.
With the increasing of $\lambda$, DRFs are trained to be more ``mature''.
This learning process is like how our human beings learn one thing from easy to hard.

\noindent\textbf{Learning $\bm{\Theta}$ and $\bm{\Pi}$.}
The parameters $\left\{\bm{\Theta},\bm{\Pi}\right\}$ and $\mathbf{v}$ are optimized alternatively.
With fixed $\mathbf{v}$,  our DRFs is optimized by alternatively learning $\bm{\Theta}$ (the parameters of DNNs) and $\bm{\Pi}$ (the parameters of regression trees).
Following~\cite{shen_deep_2018}, $\bm{\Theta}$ is optimized through gradient descent since the log of the likelihood function is differentiable with respect to $\bm{\Theta}$.
While the parameters $\bm{\Pi}$ are updated by virtue of variational bounding when fixing $\bm{\Theta}$~\cite{shen_deep_2018}.

\subsection{Prediction}
Given a new input $\mathbf{x}'$, the predictive age distribution $t'$ is then obtained through the above learned forest $\mathcal{F}$.
We take the mean of this distribution as the predicted age:

\begin{equation}
\label{Eq.6}
\hat{t}' = \int t' p_{\mathcal{F}}\left(t'|\mathbf{x}',\bm{\Theta},\bm{\Pi} \right) d t'.
\end{equation}

\begin{figure}[t]
\centerline{\includegraphics[width=8.8cm]{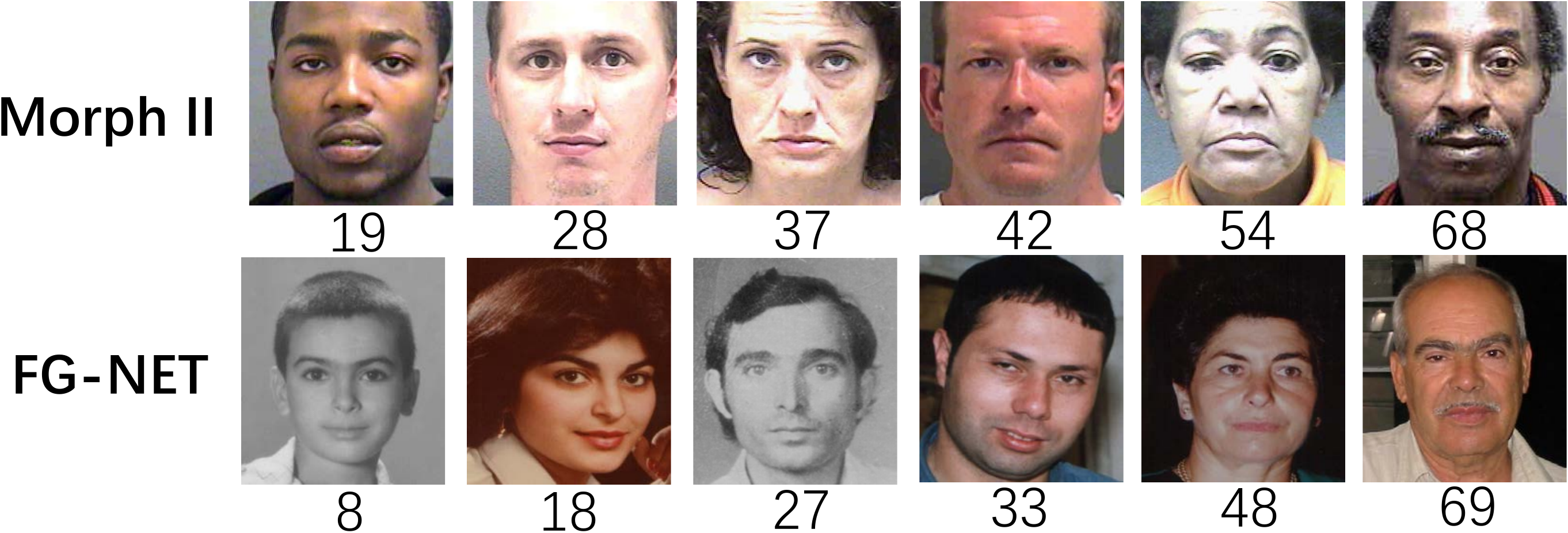}}
\caption{Some example face images in the Moprh II and FG-NET datasets. The number below each image denotes the chronological
age of that person.}
\label{pic}
\end{figure}

\section{Experiments}

We demonstrate the effecacy of SP-DRFs on two public datasets: ({\romannumeral1}) {Morph \uppercase\expandafter{\romannumeral2}~\cite{ricanek2006morph}} and ({\romannumeral2}) {FG-NET~\cite{panis2016overview}}.

\subsection{Implementation Details}

{\textbf{Datasets and configurations.}} The Morph \uppercase\expandafter{\romannumeral2} dataset contains 55,134 unique face images of 13618 individuals with unbalanced gender and ethnicity distributions.
The FG-NET dataset includes 1,002 color and grey images taken in an uncontrolled environment with huge deviations on lighting, poses and expressions of 82 people.
Following \cite{shen_deep_2018}, we randomly divided Morph \uppercase\expandafter{\romannumeral2} into two sets: 80\% of the images for training and the rest 20\% for testing.
The randomly splitting is repeated 5 times and the reported performance was obtained by averaging over these 5 five splitting.
For FG-NET, we adopted a leave-one-person-out scheme in experiments, where the images of one person are selected for testing and the remains for training.
Some example face images in these two datasets are shown in Fig.~\ref{pic}.
Our algorithm was implemented within the Caffe framework.
VGG-16~\cite{Simonyan2015} was employed as the backbone network of DRFs.
Moreover, we used the VGG-FACE~\cite{parkhi2015deep} networks as our pre-trained model.

\noindent{\bfseries Evaluation metrics.} Our methods were evaluated by~\cite{shen_deep_2018}: mean absolute error (MAE) and cumulative score (CS).
The MAE is defined as the average absolute error between the ground truth and the estimated age: $MAE=\sum_{i=1}^M\left|\widehat{t}_{i}-t_{i}\right| / M$, $\widehat{t_{i}}$ and $t_{i}$ denote the predicted and real age of the  $i^{th}$ image, and $M$ is the total number of testing images.
The CS represents the percentage of images being correctly predicted in the range of $\left[t_{i}-L, t_{i}+L\right]$: $CS(L)=\sum_{i=1}^{M}\left[\widehat{t}_{i}-t_{i} \leq L\right] /M \cdot 100 \%$, where $[\cdot]$ is an indicator function and $L$ denotes the error range.

\noindent{\bfseries Parameters setting.} The hyper-parameters of training VGG-16 were: training batch size (32 on Morph \uppercase\expandafter{\romannumeral2} and 8 for FG-NET), drop out ratio (0.5), max iterations of each retraining iteration ($80k$ on Morph \uppercase\expandafter{\romannumeral2} and $20k$ on FG-NET), stochastic gradient descent (SGD), initial learning rate (0.2 on Morph \uppercase\expandafter{\romannumeral2} and 0.02 on FG-NET) by decreasing the learning rate ($\times$0.5) per $10k$ iterations. The hyper-parameters of SP-DRFs were: tree number (5), tree depth (6), output unit number of feature learning function (128).
In addition, $\lambda$ was initialized to obtained 10\% samples in the first-round retraining for Morph
\uppercase\expandafter{\romannumeral2} dataset and 50\% for FG-NET dataset.
$\epsilon$ was set to exclude 0.5\% of the worst samples in Morph \uppercase\expandafter{\romannumeral2} and 2\% of the worst samples in FG-NET.

\noindent{\bfseries Preprocessing and data augmentation.}
We utilized multi-task cascaded CNN (MTCNN)~\cite{zhang_joint_2016} for joint face detection and alignment.
Moreover, three ways of data augmentation~\cite{shen_deep_2018} were adopted: (\romannumeral1)  clipping image with random offset, (\romannumeral2) adding Gaussian noise, and (\romannumeral3) left-right flipping.

\subsection{Validity of Self-paced Learning Strategy}
The validity of SP-DRFs was evaluated mainly on the Morph \uppercase\expandafter{\romannumeral2} dataset.
We first used all face images in training set to pre-train DRFs, and then ranked these samples on the basis of reliability (see Eq.~(\ref{Eq.3})).
Retraining started with reliable samples and gradually involved more difficult samples until it considered all samples of the training set.
After each re-training, we updated the rank of the training samples.
This process proceeded with progressively increasing $\lambda$ such that every 10\% of the whole data was gradually involved at each iteration.
Visualization of this process can be found in Fig.~\ref{Fig2}.

Fig.~\ref{Fig2} (\textbf{left}) illustrates the typical worst training samples at each iteration, along with increasing $\lambda$.
We observe that our SP-DRFs model is learned gradually to involve more difficult samples in a self-paced learning manner.
In pace 1, the most 10\% reliable samples are selected to train DRFs, even the worst cases looks normal and have relatively small prediction errors.
In the last pace, if capped likelihoods are not involved, all the samples are selected to retrain DRFs, and we observe the worst cases are face images with obvious occlusions (\eg~whiskers).
Such a visualization reveals the progressive learning process in our SP-DRFs, from simplicity to complexity.

Fig.~\ref{Fig2} (\textbf{right}) shows how self-paced learning strategy assists DRFs converges to significantly better solution.
We observe the MAE curve of SP-DRFs takes the value 3.11 at the beginning with only 10\% reliable samples being involved for retraining.
The more complex samples are involved for retraining, the better the performance changes to be.
the MAE value reduces from 3.11 to 1.99, demonstrating the leaned DRFs becomes mature when more complex samples are involved.
The reason behind this lies in that our SP-DRFs are initialized more reasonably at each pace, with more emphasis on reliable samples, and have potential ability to avoid bad local minima.

\subsection{Age Estimation Performance Comparison}

\begin{table}[t]
\centering
\renewcommand\arraystretch{1.35}
\caption{The MAE comparison on Morph \uppercase\expandafter{\romannumeral2} dataset.}\label{Table1}
\setlength{\tabcolsep}{0.5mm}
\begin{tabular}{p{2.8cm}p{1.2cm}<{\centering}p{1.4cm}<{\centering}}
\toprule
Method & MAE$\downarrow$ & CS$\uparrow$ \\
\midrule\midrule
CCA \cite{guo2013joint} & 4.73 & 60.5\%*\tablefootnote{* denotes the result is from the reported CS curve.} \\
LSVR \cite{guo_human_2009} & 4.31 & 66.2\%* \\
RCCA \cite{Huerta2014Facial} & 4.25 & 71.2\% \\
OHRank \cite{Chang2011Ordinal} & 3.82 & N/A \\
OR-CNN \cite{niu_ordinal_2016} & 3.27 &	73.0\%* \\
Ranking-CNN \cite{chen_using_2017} & 2.96 & 85.0\%* \\
DEX \cite{rothe2018deep} & 2.68 & N/A \\
DLDL \cite{gao2017deep} & 2.42 & N/A \\
dLDLF \cite{Shen2017Label} & 2.24 & N/A \\
DRFs \cite{shen_deep_2018} & 2.17 & 91.3\% \\
\textbf{SP-DRFs} & \textbf{1.99} & \textbf{92.78\%} \\
\textbf{SP-DRFs (capped)} & \textbf{1.98} & \textbf{92.79\%} \\
\bottomrule
\end{tabular}
\end{table}
\begin{figure}[t]
\centering
\includegraphics[width=0.48\textwidth]{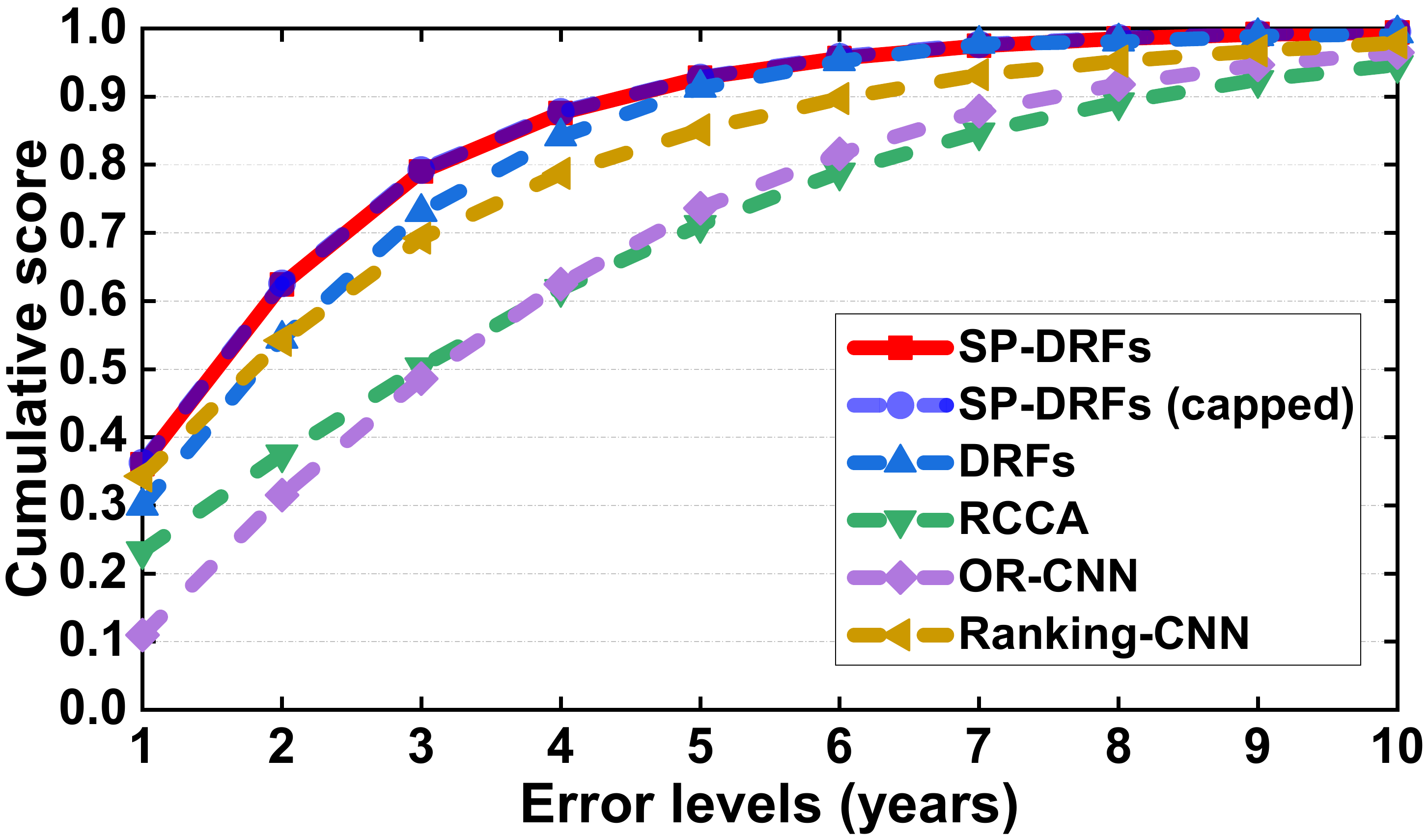}
\caption{The CS comparison on the Morph II dataset}
\label{Fig3}
\end{figure}

We compared SP-DRFs with the state-of-the-art age estimation approaches on both Morph \uppercase\expandafter{\romannumeral2} and FG-NET datasets.

\noindent\textbf{Morph \uppercase\expandafter{\romannumeral2}.} Tab.~\ref{Table1} shows the comparison results on the Morph \uppercase\expandafter{\romannumeral2} dataset, where some consistent trends are found.
First, the SP-DRFs significantly outperform shallow model based approaches, such as LSVR~\cite{guo_human_2009} and OHRank~\cite{Chang2011Ordinal}, due to its high model capacity.
Second, the SP-DRFs outperform the existing DNNs based methods by achieving a 1.99 MAE.
Importantly, comparing with DRFs, SP-DRFs promote the MAE by 0.19, demonstrating the validity of the gradual learning mechanism in SP-DRFs.
This is owing to SP-DRFs' ability to progressively retrain DRFs and hence suppress the interference arising from confusing and noisy samples, and decrease the risk of stucking in bad local optima.
Third, after introducing capped likelihood in the SP-DRFs, the noisy samples are excluded, leading to more improved performance.


Fig.~\ref{Fig3} shows the CS comparison on the Morph \uppercase\expandafter{\romannumeral2} dataset.
We observe SP-DRFs' CS achieves 92.78\% at the error level $L=5$, which obviously outperform DRFs and obtain a 1.48\% increment.
Also, SP-DRFs and the capped-likelihood SP-DRFs have the highest CS at every error level.
Importantly, comparing with the CS curve of DRFs, SP-DRFs have obvious performance improvement, which demonstrates the effect of introducing gradual learning mechanism again.

\begin{table}[t]
\centering
\renewcommand\arraystretch{1.35}
\caption{The MAE comparison on the FG-NET dataset.}\label{Table2}
\setlength{\tabcolsep}{0.5mm}
\begin{tabular}{p{2.8cm}p{1.2cm}<{\centering}p{1.4cm}<{\centering}}
 \toprule
Method & MAE$\downarrow$ & CS$\uparrow$ \\
\midrule\midrule
AGES \cite{geng2007automatic} & 6.77 & 64.1\%* \\
IIS-LDL \cite{xin_geng_facial_2013} & 5.77 &	N/A \\
LARR \cite{guodong_guo_image-based_2008} & 5.07 & 68.9\%* \\
MTWGP \cite{Yu2010Multi} & 4.83 & 72.3\%* \\
DIF \cite{han_demographic_2015} & 4.80 & 74.3\%* \\
CA-SVR \cite{chen2013cumulative} & 4.67 & 74.5 \\
OHRank \cite{Chang2011Ordinal} & 4.48 & 74.4\% \\
DLA \cite{wang2015deeply} & 4.26 & N/A \\
CAM \cite{Luu2013Contourlet} & 4.12 & 73.5\%* \\
DRFs \cite{shen_deep_2018} & 3.85 & 80.6\% \\
\textbf{SP-DRFs} & \textbf{3.46} & \textbf{81.44\%} \\
\textbf{SP-DRFs (capped)} & \textbf{3.44} & \textbf{81.84\%} \\
\bottomrule
\end{tabular}
\end{table}
\begin{figure}[t]
\centering
\includegraphics[width=0.48\textwidth]{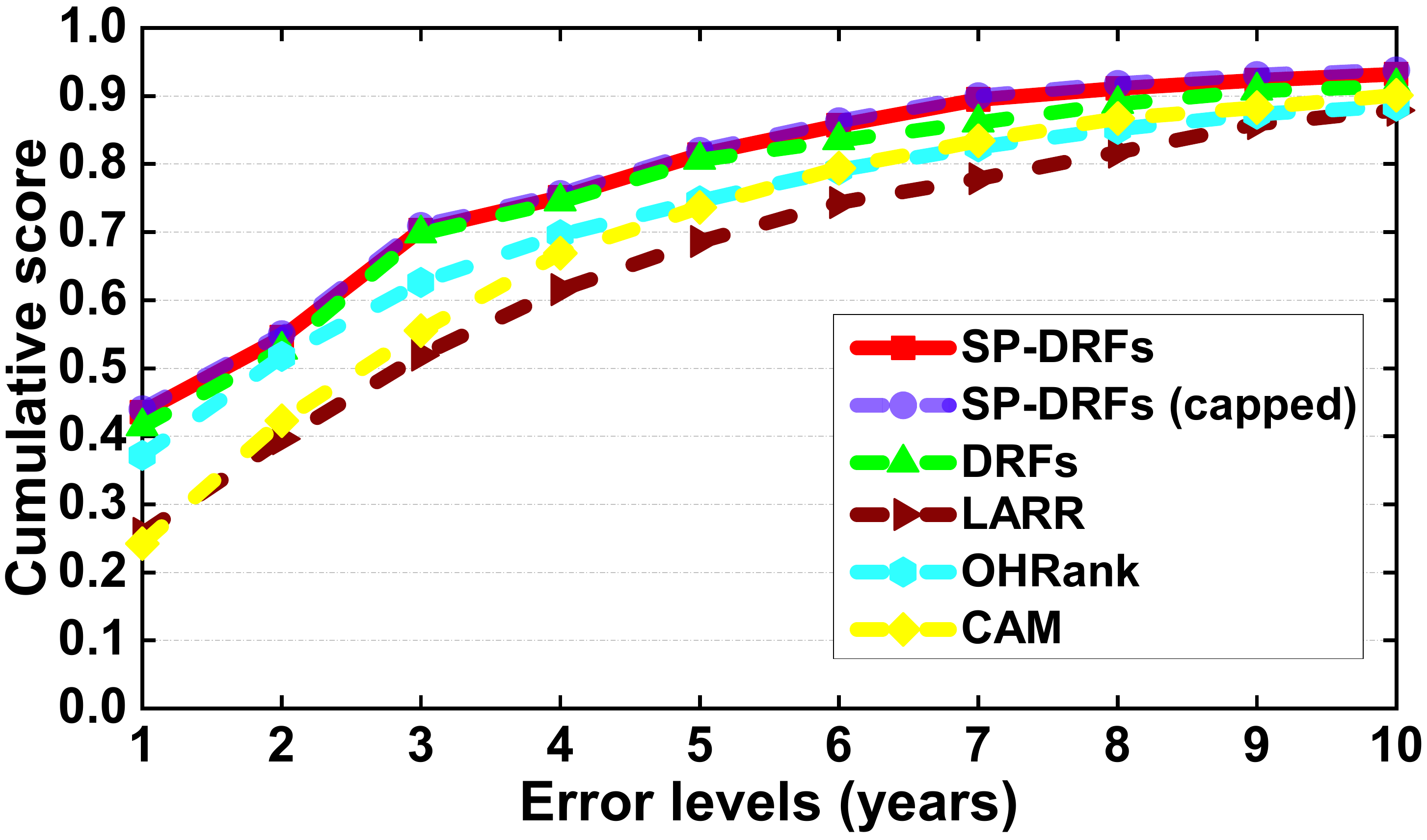}
\caption{The CS comparison on the FG-NET dataset}
\label{Fig4}
\end{figure}

\noindent\textbf{FG-NET.}
Tab.~\ref{Table2} shows the MAE comparison of SP-DRFs with other state-of-the-art age estimation methods.
As can be seen, SP-DRFs achieves an MAE of 3.46 and the capped-likelihood SP-DRFs even reaches an MAE of 3.44, which reduces the performance of DRFs by 0.41 years.
As we discussed previously, this is owe to the gradual learning strategy which improves the robustness of DRFs.
Besides, introducing capped likelihood slightly promotes both the MAE and CS (error level $L=5$).

Fig.~\ref{Fig4} shows the CS comparison with recent proposed age estimation methods at different error levels on the FG-NET dataset.
The SP-DRFs and the capped-likelihood SP-DRFs both consistently outperform other methods at different error levels and exhibit improved robust performance.
Compared with DRFs, SP-DRFs also show obvious superiority.
The results, once again, demonstrate the efficacy of the gradual learning mechanism in our SP-DRFs on improving the robustness of facial age estimation.

\section{Conclusion}

We proposed a novel facial age estimation approach, namely self-paced deep regression forests (SP-DRFs).
Through progressively selecting the training samples from easy to hard, our SP-DRFs can be trained iteratively to obtain better solution -- not only robust to noisy but also away from bad local minima.
Meanwhile, we proposed to construct capped likelihood function in our SP-DRFs to further exclude noisy samples and achieved more robust results.
Experiments on well-known facial age estimation datasets demonstrated the SP-DRFs achieve the state-of-the-art performance.
The future work will include exploring how to combine self-paced leaning paradigm with other facial age estimation methods.

\section*{Acknowledgment}
This work is supported by the National Natural Science Foundation of China (Nos. 61806043, 61971106, 61872068 and 61603077), the China Postdoctoral Science Foundation (Nos. 2016M602674 and 2017M623007), and the grant from Science and Technology Department of Sichuan Province of China (Nos. 2018GZ0071 and 2019YFG0426).

\normalsize
\bibliographystyle{IEEEtran}
\bibliography{reference}

\end{document}